\documentclass[runningheads]{llncs}
\usepackage{graphicx}

\usepackage{tikz}
\usepackage{comment} 
\usepackage{amsmath,amssymb} 
\usepackage{color}
\usepackage{epsfig}
\usepackage{multirow}
\usepackage{threeparttable}
\usepackage{array}

\newcommand{\wha}{\textcolor[rgb]{0,0.0,0.0}}
\newcommand{\whb}{\textcolor[rgb]{0,0.0,0.0}}
\newcommand{\whc}{\textcolor[rgb]{0.0,0.0,0.0}}
\newcommand{\whd}{\textcolor[rgb]{0,0.0,0.0}}
\newcommand{\whe}{\textcolor[rgb]{0,0.0,0.0}}

\begin{document}
\pagestyle{headings}
\mainmatter
\def\ECCVSubNumber{2718}  

\title{Contextual Heterogeneous Graph Network for Human-Object Interaction Detection} 

\titlerunning{Contextual Heterogeneous Graph Network for HOI}
%
\author{Hai Wang\inst{1,2} \and
Wei-shi Zheng\thanks{Corresponding author}\inst{1,2,3} \and
Ling Yingbiao\inst{1}}
\authorrunning{H. Wang, W.-S. Zheng, L. Yingbiao}
%
\institute{\footnotesize School of Data and Computer Science, Sun Yat-sen University, China \and
Key Laboratory of Machine Intelligence and Advanced Computing, Ministry of Education, China \and
Peng Cheng Laboratory, Shenzhen 518005, China\\
\email{wangh577@mail2.sysu.edu.cn \quad wszheng@ieee.org \quad isslyb@mail.sysu.edu.cn}}
\maketitle

\begin{abstract}
Human-object interaction (HOI) detection is an important task for understanding human activity. Graph structure is appropriate to denote the HOIs in the scene. Since there is an subordination between human and object---human play subjective role and object play objective role in HOI, 
the relations between homogeneous entities and heterogeneous entities in the scene should also not be equally the same. However, previous graph models regard human and object as the same kind of nodes and do not consider that the messages are not equally the same between different entities. In this work, we address such a problem for HOI task by proposing a heterogeneous graph network that models humans and objects as different kinds of nodes and incorporates intra-class messages between homogeneous nodes and inter-class messages between heterogeneous nodes.
In addition, a graph attention mechanism based on the intra-class context and inter-class context is exploited to improve the learning. 
Extensive experiments on the benchmark datasets V-COCO and HICO-DET verify the effectiveness of our method and demonstrate the importance to extract intra-class and inter-class messages which are not equally the same in HOI detection.
\end{abstract}

\section{Introduction}

Given a still image, human-object interaction (HOI) detection requires detecting the positions of people and objects and reasoning about their interactions, such as ``eating pizza" and ``drinking water". HOI is crucial for understanding the visual relationships \cite{lu2016visual,krishna2017visual} between entities in human activity and has attracted increasing interest. Over the past few years, tasks of recognizing individual visual objects, \emph{e.g.}, object detection \cite{girshick2015fast,lin2017feature} and pose estimation \cite{cao2017realtime,chu2017multi}, have witnessed impressive progress thanks to the development of deep learning. Inspired by this progress, a few deep learning methods have been proposed to address HOI detection tasks.

\begin{figure}[t]
\centering
\includegraphics[width=\linewidth]{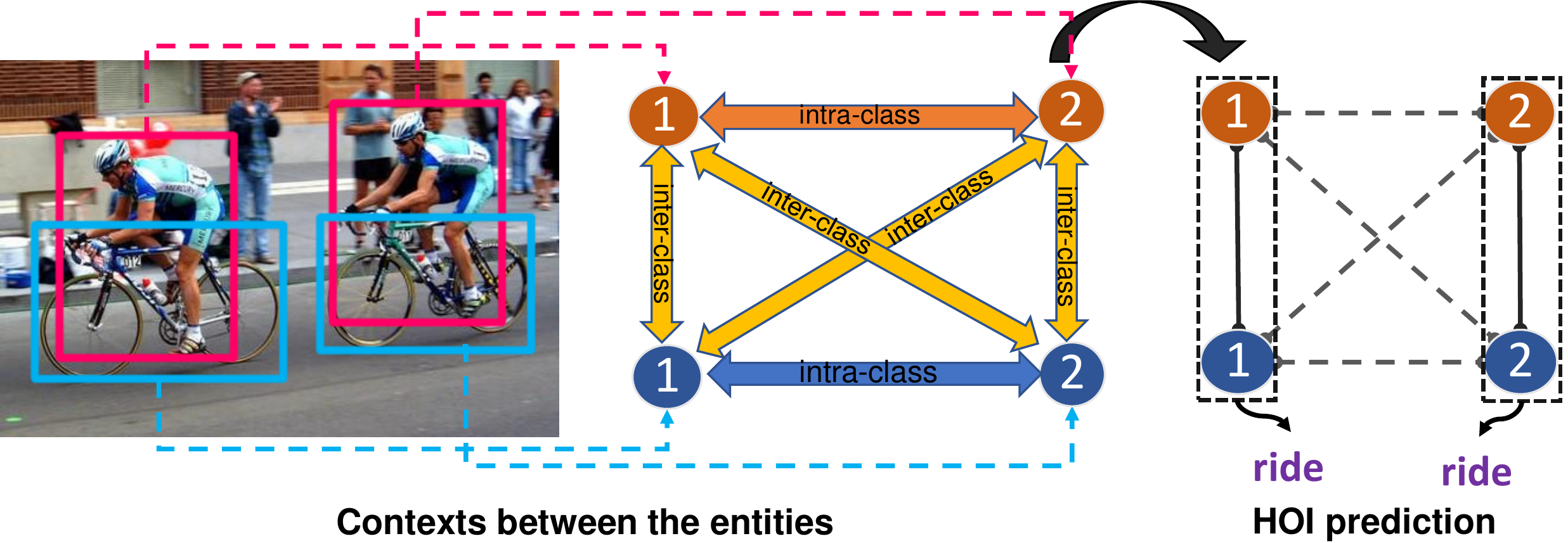} 
\caption{In the picture on the left, we can see that the two people are subjective in the interaction of ``riding" with their bicycles which are objective. The intra-class context between two riders or bicycles is the similarities in some aspects reflecting HOI and the inter-class context between the rider and bicycle is the interactiveness (whether they have interaction). The contexts can be explored to improve HOI recognition in the picture on the right.} 
\label{intro}
\end{figure}

HOI recognition considers not only the human, object instances, but also the relations of the entities. Many works \cite{prest2011weakly,yao2010modeling,yao2010grouplet,wan2019pose,wang2019deep} have studied the interaction of the human-object pair utilizing spatial context or fine-grained features. Generally, an activity scene includes a number of humans and objects. However, most works only focused on recognizing the interaction(s) of one human-object pair and neglect scene understanding. Since multiple relations potentially exist in the complicated scene, the relationship reasoning is important for a more comprehensive understanding of human activity. \cite{qi2018learning} employed a graph model to infer the relationships, while the authors modeled the people and objects as the same kind of nodes and all the relations as the same edges in the graph which equally treats all the relations in activity scene as interactions. However, human and object play different roles (subject and object) in HOI. There are inter-class context between heterogeneous entities (subject and object) and intra-class context between homogeneous entities (subject-subject, object-object) in activity scene, which means the relations are not equally the same. For example, in Fig. \ref{intro}, the inter-class context between the rider and bicycle is interactiveness which means whether they are interacting with each other, and the intra-class context between riders (bicycles) can be the similarities in the aspects related to HOI, like appearance, pose or spatial configurations, which are different with inter-class context. Therefore, processing all the relations in a same way is defective. If we further explore messages from the unequal contexts in the scene to better process the relationships, HOI detection can be improved.

Since it is convenient to represent the entities as nodes and the relations as the edges connecting them, the graph structure is appropriate for modeling HOI. In this paper, we propose our contextual heterogeneous graph network to deeper explore the relations between humans and objects. We discriminatively represent human and object entities as heterogeneous nodes and elaborate different propagating functions for the message passing between heterogeneous nodes and homogeneous nodes. 
The spatial relation of a person and an object is essential information for recognizing the interaction and is encoded into the edges that connect heterogeneous nodes. The edges connecting homogeneous nodes represent the intra-class context, which reflects the relevance of homogeneous nodes, while the edges connecting heterogeneous nodes represent the inter-class context which reflects interactiveness. In addition, we combine the contextual learning with the graph attention method \cite{velivckovic2017graph} to improve the effectiveness when the nodes gather knowledge from their neighbors.  

Our contributions can be summarized as follows. Recognizing the characteristics of HOI, we (1) explore the unequal relations and messages which are not studied before between humans and objects (subjective and objective roles) in activity scenes and (2) propose a graph-structure framework with heterogeneous nodes and attention mechanism to learn the intra-class and inter-class messages.
(3) We evaluate our method on two evaluation datasets, V-COCO and HICO-DET, and we find that our heterogeneous node graph model is effective and outperforms state-of-the-art methods. 

\section{Related Work}

\noindent\textbf{Human-object interaction detection}
is important for understanding human activity in social scenes. 
Studies performed before the advent of deep learning produced some interesting conclusions. \cite{gupta2009observing} used the Bayesian model to study interactions. Given the localization of a human and object, \cite{prest2011weakly} learned the probability distribution of their spatial relationship. A representation for inferring the spatial relationship from the visual appearance reference point was proposed by \cite{yao2010grouplet}. 
Traditional methods also include utilizing atomic pose exemplars \cite{hu2013recognising}, learning a compositional model \cite{desai2010discriminative}, and exploiting relationship of body parts and objects \cite{delaitre2011learning,yao2010modeling}. 

Several HOI datasets, such as V-COCO \cite{gupta2015visual} and HICO \cite{chao2015hico}, have been reported over the years. With the assistance of DNN, the methods proposed over the years have achieved significant progresses. Some zero-shot learning methods have gained great progresses. \cite{shen2018scaling} addressed the issue by predicting verbs in images and \cite{peyre2019detecting} recognized the interactions by the analogies between the pair-wise relations. \cite{chao2018learning} utilized a multistream network to learn human and object features and the corresponding spatial relations separately.
In \cite{gkioxari2018detecting}, a network that utilizes human action to estimate the object localization density map was proposed. \cite{gao2018ican} proposed an instance attention module to highlight the instance region. Pose information was exploited in works like the turbo learning framework \cite{feng2019turbo} and the fine-grained multi-level feature network \cite{wang2019deep}, which have made impressive results. Fine-grained methods also include RPNN \cite{zhou2019relation} which designed the bodypart graphs to imrove localization of related objects and action recognition. In \cite{li2018transferable}, the authors focused on suppressing noninteracting HOI pairs in images before recognition by a two-stage model. \cite{wang2019deep} used an attention module to select instance-centric context information. We further infer the HOI relationships by learning the unequal messages between different entities and propose a graph network which denotes the subjective entities and objective entities as heterogeneous nodes.

\noindent \textbf{Graph neural networks (GNNs)}
combine a graphical model and a neural network. A GNN learns potential features for nodes by iteratively propagating messages from neighboring nodes and updating hidden states by embedded functions. \cite{scarselli2008computational} provided a formal definition of early GNN approaches and demonstrated the approximation properties and computational capabilities of GNNs. \cite{kipf2016semi} constructed a graph convolution network (GCN) to learn a scalable graph-structure context. In \cite{gilmer2017neural}, the MPNN framework was proposed as a general model that abstracts the commonalities of several well-known graph networks.
\cite{velivckovic2017graph} combined the attention mechanism with a GNN and demonstrated their graph attention network (GAT). GraphSAGE \cite{hamilton2017inductive} aggregates neighbor nodes, propagates to the far nodes layer by layer, and generates embeddings for unseen nodes by iterative propagations. 
Visual classification was processed by constructing knowledge graph \cite{lee2018multi,wang2018zero} and skeleton-based action learning task was also improved by more comprehensive pose information learned by graph structure \cite{li2019actional}. Heterogeneous graph has ever been exploited in \cite{yu2019heterogeneous} to address VQA task, while their graph infers a global category and ours infers the relationships of heterogeneous nodes. 

The graph models mentioned above have achieved significant results in their specific fields. For HOI detection, GPNN \cite{qi2018learning} applied a GNN which is a homogeneous graph and all instances were treated as the same kind of nodes. We notice that the contexts between different nodes could be unequal;thus, we further design a heterogeneous graph to infer the interaction. The details of the heterogeneous graph network is presented in the next section. 

\begin{figure}[t]
    \centering
    \includegraphics[width=0.9\linewidth]{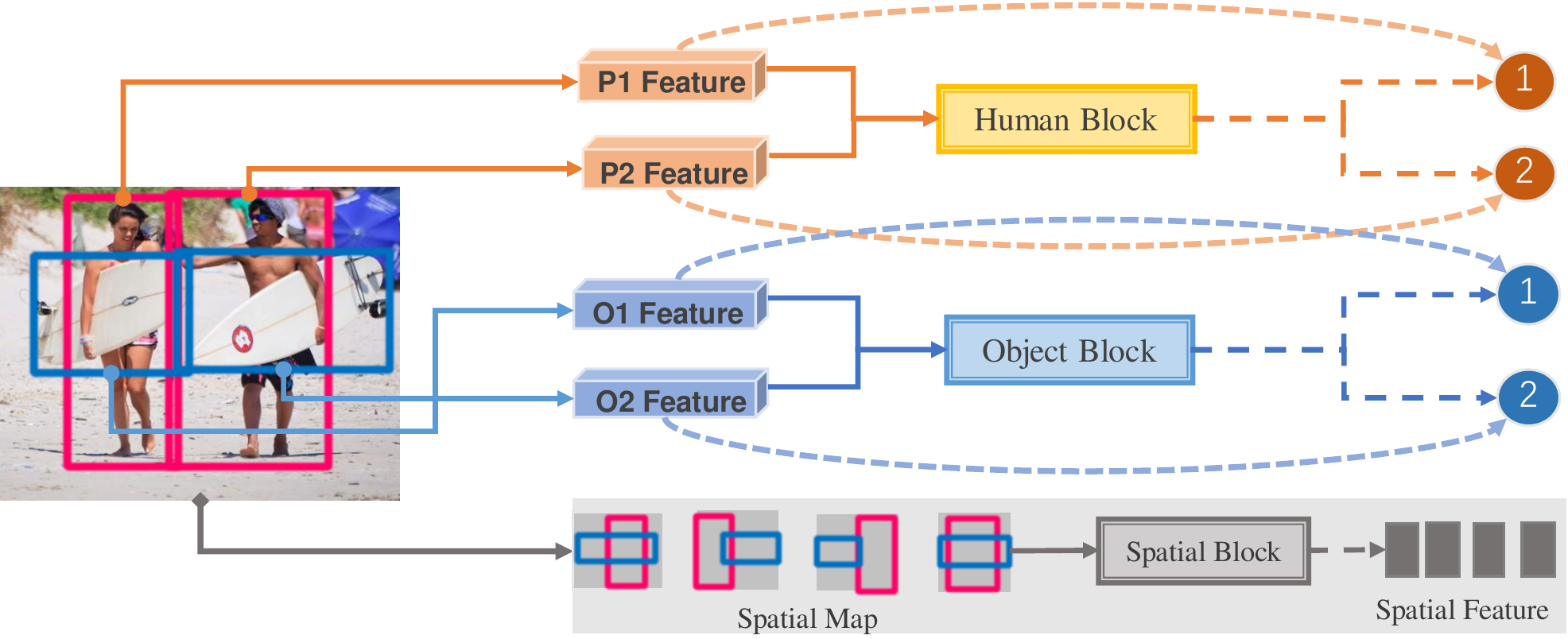} 
    \caption{An overview of the feature extraction module. The spatial map is introduced in \ref{cl}. The human block and object block further extract the nodes' initial states from the proposal region features. After extraction, the people in the image are denoted as circles in deep orange, and the objects(surfboards) are similarly denoted as the circles in light blue.}
    \label{extract}
\end{figure}

\section{Approach}

\subsection{Preliminary}
Given an image, the HOI task is to first detect the people and objects and then predict the interaction label of every human-object pair. We denote the people who play subjective roles as \wha{subject nodes} {$p_1, p_2,\cdots,p_N$}, and objects as object nodes {$o_1, o_2,\cdots,o_M$} in our graph. The hidden embedding of the \wha{subject node} is denoted as $\mathbf{h}_p$ and the object node as $\mathbf{h}_o$. 
In the following subsections, we introduce the pipeline of our heterogeneous graph network, the intra-class and inter-class context learning functions and adopted graph attention method, prediction function, sequentially.

\whe{\subsection{Pipeline}}
Fig. \ref{extract} shows an overview of the feature extraction module. After detection, we extract the human and object features, as well as the spatial configurations of the human-object pairs by the multistream extraction module with the bounding boxes produced by the object detector \cite{ren2015faster}. HOI representation can be denoted as a graph $\mathcal{G}(\mathcal{V}, \mathcal{E})$, where human and object entities are represented as nodes in node set $\mathcal{V}$ and the relations among the entities are denoted as edges in edge set $\mathcal{E}$. Our heterogeneous graph is a fully connected graph consisting of two types of nodes, \wha{subject nodes} and object nodes, which belong to two subsets, $\mathcal{V}_p$ and $\mathcal{V}_o$, respectively. The initial states of the nodes are the extracted features of the entities. 
The characteristics of the relations among homogeneous entities and heterogeneous entities are not equally the same: the former represents the similarity in some aspects which relate to HOI between homogeneous entities, while the latter represents the interactiveness of a subject and an object. We regard them as intra-class context and inter-class context respectively, and are denoted as two kinds of edges in the heterogeneous graph. 

\begin{figure}[t]
    \centering
    \includegraphics[width=\linewidth]{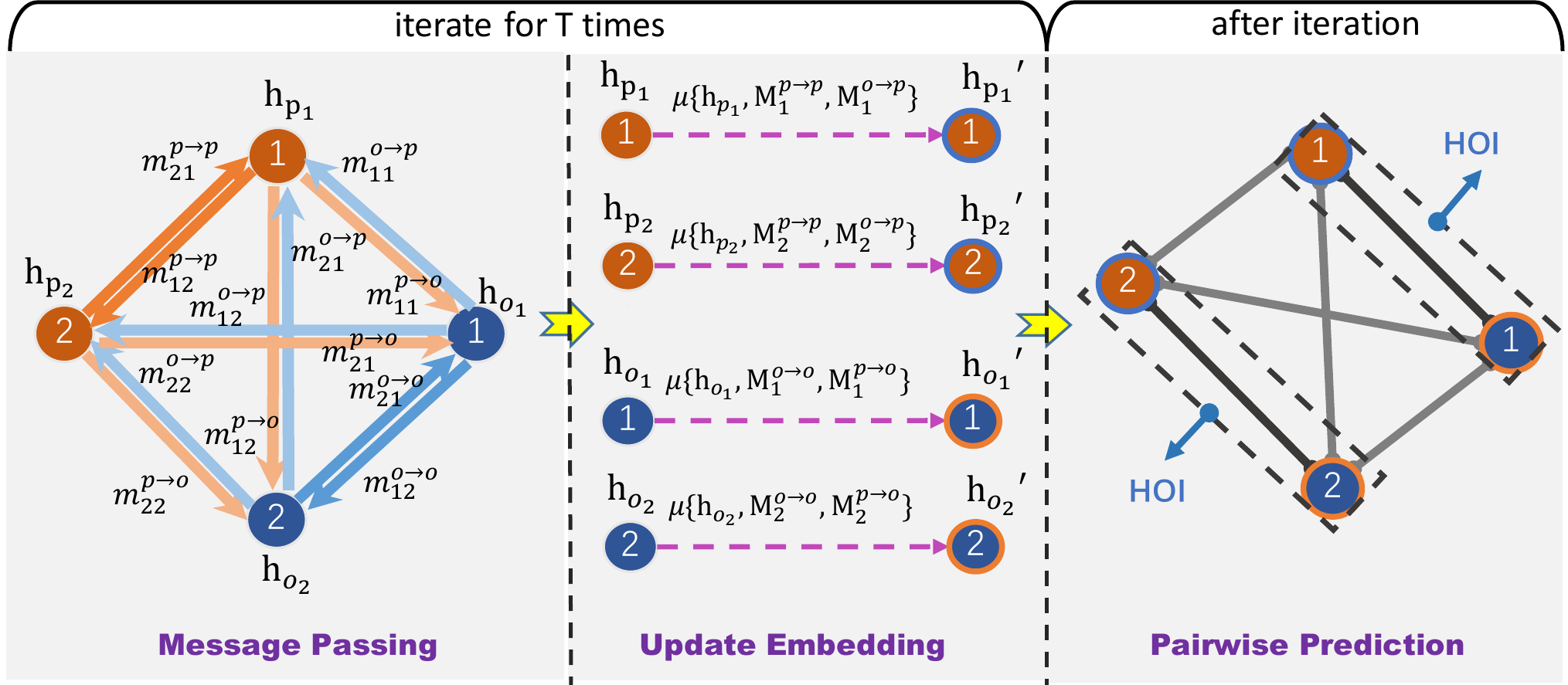} 
    \caption{An illustration of the reasoning procedure of our graph neural network. Each node gathers messages from homogeneous neighborhoods ($m^{p\rightarrow p}$ or $m^{o\rightarrow o}$) and heterogeneous neighborhoods ($m^{p\rightarrow o}$ or $m^{o\rightarrow p}$) in a fully connected graph. The messages from homogeneous nodes contain intra-class information that strengthens the expression, and the messages from heterogeneous nodes contain inter-class information about what interactions they have and who they interact with. Each node updates the hidden state with the previous state and the gathered messages by update function $\mu(\cdot)$. Finally the model predicts the label of each HOI pair. }
    \label{framework}
\end{figure} 

The relationship reasoning procedure is illustrated in Fig. \ref{framework}. Our learning strategy is to iteratively propagate messages between nodes to gather information from other nodes and update the node hidden embeddings. The first stage of the relationship reasoning procedure is the message passing stage, which consists of two steps: intra-class message passing between homogeneous nodes and inter-class message passing between heterogeneous nodes. For each node ${v}\in\mathcal{V}$, the intra-class message aggregation method can simply be formulated as follows:
\begin{equation}\label{eq:homo}
    \whe{\mathbf{M}}_v^{intra} = \whe{\mathrm{Agg}_1}(\{\whe{\mathcal{F}_1}(\mathbf{h}_u,\ \whe{\mathbf{e}_{vu}}),\quad \forall u\in \mathcal{N}_v^{intra}\}),\\
\end{equation}
where $\mathbf{M}_v^{intra}$ (includes $m^{p\rightarrow p}$ for \wha{subject nodes} and $m^{o\rightarrow o}$ for object nodes in Fig. \ref{framework}) is the intra-class message that $v$ gathers from its homogeneous neighborhood $\mathcal{N}_v^{intra}$ and \whe{$\mathbf{e}_{vu}$} is the edge that connects $v$ and neighbor node $u$, thus representing the intra-class context among them. The aggregation function \whe{$\mathrm{Agg}_1(\cdot)$} and message passing function \whe{$\mathcal{F}_1(\cdot)$} in Eq.\ref{eq:homo} will be clearly presented in subsection \ref{cl}. The intuition behind Eq.\ref{eq:homo} can be interpreted as follows: an entity can learn the information about the interaction from relevant entities that have the same or similar interaction by the graph deep reasoning. Similarly, the inter-class meassge aggregation can be denoted by the following: 
\begin{equation}\label{eq:diver}
    \whe{\mathbf{M}}_v^{inter} = \whe{\mathrm{Agg}_2}(\{\whe{\mathcal{F}_2}(\mathbf{h}_u,\ \whe{\mathbf{e}_{vu}}),\quad \forall u\in \mathcal{N}_v^{inter}\}),\\
\end{equation}
where $\mathbf{M}_v^{inter}$ (includes $m^{o\rightarrow p}$ for \wha{subject nodes} and $m^{p\rightarrow o}$ for object nodes in Fig. \ref{framework}) represents the inter-class message propagated from heterogeneous neighborhood of $v$. The idea of Eq.\ref{eq:diver} can be interpreted as follows: the features of two heterogeneous entities which interact with each other should have some consistency with respect to the semantic knowledge of interaction after some learning processes. For example, we know that a tennis racket is used to play tennis and that a man waving his arm forward above his head is likely to bat and thus both the racket and the man are related to ``play tennis" to some extent. From this perspective, they are consistent in ``playing tennis" and supposed to learn from each other, by which they could ``know each other better" and strengthen their interaction relation.
We demonstrate the explicit information aggregation strategies in Eq.\ref{eq:homo} and Eq.\ref{eq:diver} in the next subsection after introducing intra-class context and inter-class context.

With the message learned in the first stage, we update the hidden embedding of node $v$ in the next stage:
\whe{\begin{equation}\label{eq:update}
    \mathbf{h'}_v = \mu(\mathbf{h}_v + \mathbf{M}_v^{intra} + \mathbf{M}_v^{inter}) + \mathbf{h}_v^0,
\end{equation}}
where $\mathbf{h}_v^0$ is the initial node embedding and $\mathbf{h'}_v$ is the node embedding in the next iteration time after the update. Our update function can be realized as an MLP such as a fully connected layer with an activation, denoted as function $\mu$ in Eq.\ref{eq:update}. 

After the graph reasoning process, we predict the interaction label of every HOI pair by a joint classifier. We use $f_{cls}(\cdot)$ to represent the classifier which realized by an FC layer with sigmoid function. The prediction $\mathbf{y}_{ij}$ of the pair is computed as follow:
\begin{equation}\label{eq:pred}
    \mathbf{y}_{ij} = f_{cls}(\mathbf{h}_{p_i}+\mathbf{h}_{o_j}+\whe{\mathbf{e}_{p_io_j}}).
\end{equation}

\subsection{Contextual Learning}\label{cl}

\noindent \textbf{Intra-class Learning.}
\whb{Generally, human activity scene is a complicated scene which involves multiple entities, so contextual learning is important to infer the relationships between humans and objects. In our heterogeneous graph, the intra-class context is the similarity between homogeneous nodes with respect to the interactions.} It is intuitive that if many people are interacting with the same or similar object(s), \emph{e.g.}, a group of people are rowing a boat, they are very likely performing the same or relevant actions. 
In order to explore the intra-class context, \wha{for each node we compute a vector $\mathbf{r}$, which will be used to compute intra-class context.} We denote $f(\cdot)$ as the function realized by an FC layer with ReLU. The computation of vector $\mathbf{r_i}$ of node $p_i$ is formulated as follows:
\begin{equation}
    \mathbf{r}_i = \mathbf{max}\{f(\mathbf{h}_{p_i}+\mathbf{h}_{o_j}+\mathbf{s}_{p_io_j}),\quad \forall o_j\in \mathcal{N}_{p_i}^{inter}\},
\end{equation}
where $\mathbf{s}_{p_io_j}$ is the encoded spatial configuration feature between $p_i$ and $o_j$. To extract the spatial feature, we construct a 2-channel spatial map consisting of a human map and an object map for a human-object pair. Each map has one channel with a value of 1 in the area of the human(object) position and 0 in others. Finally, we resize the scale to 64$\times$64. a spatial location encoder which consists of CNN layers and FC layers will be used to extract the spatial feature from the map. 
Here, $\mathbf{r}_i$ can reflect the knowledge about the interactions $p_i$ have in this scene, since $\mathbf{h}_{p_i}$ contains information about the appearance of $p_i$, and the combination of $\mathbf{h}_{p_i}$, $\mathbf{h}_{o_j}$ and $\mathbf{s}_{p_io_j}$ contains the information for inferring the interaction between $p_i$ and $o_j$. 

The contextual learning are utilized with graph attention in message passing layers. For nodes $p_i$ and $p_j$, we first compute the cosine similarity of $\mathbf{r}_i$ and $\mathbf{r}_j$ to represent the intra-class context between them:
\begin{equation}\label{eq:eps}
    \epsilon_{ij} = \frac{\overrightarrow{\mathbf{r}_i}\cdot\overrightarrow{\mathbf{r}_j}}{\left\|\overrightarrow{\mathbf{r}_i}\right\|\left\|\overrightarrow{\mathbf{r}_j}\right\|},
\end{equation}
The attention weight in the process of gathering homogeneous nodes' messages is their normalization after the softmax function:
\begin{equation}\label{eq:alpha}
    \alpha_{ij} = \frac{\mathrm{exp}(\epsilon_{ij})}{\sum_{k=1}^M\mathrm{exp}(\epsilon_{ik})}.
\end{equation}
In summary, to node $p_i$, Eq.\ref{eq:homo} can be clearly rewritten as the following:
\begin{equation}\label{eq:homo_2}
    \mathbf{M}_{p_i}^{intra} = \mathrm{mean}(\sum_{p_j\in \mathcal{N}_{p_i}^{intra}}\alpha_{ij}\cdot f(\mathbf{h}_{p_j})).
\end{equation}
The computation of $\mathbf{r}$ for object nodes is similar, and the graph attention of object nodes is exploited in the same way. \whc{By learning with the intra-class context, the nodes will be able to tell which homogeneous nodes are relevant and gather more messages from them, while the irrelevant nodes will be eliminated during the reasoning. }

\noindent \textbf{Inter-class Learning.}
The inter-class context is the interactiveness that reflects prior interactive knowledge in HOI. We introduces a interactiveness weight $\mathbf{w}$ to represent the inter-class context. Inspired by GPNN \cite{qi2018learning}, the interactiveness weight of \wha{subject node} ${p}_i$ and object node ${o}_j$ is computed by an FC layer with activation:
\begin{equation}
    \mathbf{w}_{ij} = f(\mathbf{h}_{p_i}+\mathbf{h}_{o_j}+\mathbf{s}_{p_io_j}),\qquad {w}_{ij}\in\mathbb{R}.
\end{equation}
During training, we minimize the binary classification cross-entropy loss $\mathcal{L}_w$ to optimize \whe{$\mathbf{w}$}:
\begin{equation}
\begin{split}
    \mathcal{L}_w = &-\frac{1}{NM}\sum_{i}^{N}\sum_{j}^{M}(\mathbf{1}_{ij}\cdot log(\hat{\mathbf{w}}_{ij})\\
    &+(1-\mathbf{1}_{ij})\cdot log(1-\hat{\mathbf{w}}_{ij})),
\end{split}
\end{equation}
where the label is 1 if the human and object are interacting with each other and 0 otherwise. ${N}$ and ${M}$ are the numbers of people and objects. The loss function is used to learn whether the subject node ${p}_i$ is interacting with the object node ${o}_j$. 

The inter-class message that $p_i$ gets from $o_j$ is a combined knowledge of the hidden embedding $\mathbf{h_{o_j}}$ and the spatial configuration $\mathbf{s}_{p_io_j}$ because they are both necessary for recognizing the pairwise interaction. Adopting graph attention mechanism, to node $p_i$ Eq.\ref{eq:diver} can similarly be rewritten as:
\begin{equation}\label{eq:diver_2}
    \begin{split}
        \mathbf{M}_{p_i}^{inter} = &\mathrm{max}\big\{\frac{\mathrm{exp}(\mathbf{w}_{ij})}{\sum_{k=1}^M\mathrm{exp}(\mathbf{w}_{ik})} 
        \cdot f(\mathbf{s}_{p_io_j}\oplus\mathbf{h}_{o_j}),\quad \forall o_j\in \mathcal{N}_{p_i}^{inter}\big\}, 
    \end{split}
\end{equation}
where $\oplus$ means concatenation. Since inter-class context is expected to tell whether the entities have interaction or not, the maximization can help the inter-class message focus on those interactive heterogeneous nodes and neglect many non-interactive nodes. The inter-class learning strengthens the relations between heterogeneous nodes which are interactive. 


\subsection{HOI Prediction}
After the reasoning of the graph, the model predicts the interaction of each HOI pair by a \wha{joint classifier}. Our prediction is computed as follows:
\begin{equation}\label{eq:pred_2}
    \mathbf{y}_{ij} = f(\mathbf{h}_{p_i}+\mathbf{h}_{o_j}+\mathbf{s}_{p_io_j}).
\end{equation}
During training, We minimize the binary classification cross-entropy loss $\mathcal{L}_{ho}$ for prediction. In summary, the objective function is the weighted sum of the interaction classification loss $\mathcal{L}_{ho}$ and interactiveness loss $\mathcal{L}_w$:
\begin{equation}\label{loss}
    \mathcal{L} = \lambda \cdot \mathcal{L}_{ho} + \mathcal{L}_w.
\end{equation}
During inference, we compute the HOI detection score $s$ of $h$ and $o$ with the prediction score $\mathbf{y}$:
\begin{equation}\label{score}
    s = \mathbf{y} \cdot s_h \cdot s_o,
\end{equation}
\whc{where} $s_h$, $s_o$ are the object detection confidences of $h$ and $o$. 

\section{Experiments}

\subsection{Datasets and Metrics}
\noindent\textbf{Datasets. }
We evaluate our model on two HOI datasets: V-COCO and HICO-DET. \textbf{V-COCO} \cite{gupta2015visual} is a subset of COCO2014 \cite{lin2014microsoft} that includes 5,400 images, 8431 people instances in trainval set and 4946 images, 7768 people instances in test set. The instance-level annotations are for the classification of 29 action categories (5 of them are only for humans). In addition, there are two roles of interaction targets in the annotations: instrument and direct object, which should be classified in recognition.

\noindent\textbf{HICO-DET}
\cite{chao2018learning} includes the image-level classes' annotations of HICO and further provides instance-level bounding boxes for instances and pair-wise class labels. There are 38,118 images and 117,871 instances in the training subset, with 9,658 images, and 33,405 instances in the testing subset. The annotations consist of 600 HOI categories and 117 action categories.

\noindent\textbf{Metric. }
We adopt the mean average precision (mAP), which is generally used in detection tasks, for our evaluation. We regard a prediction as a true positive when both the human and object bounding boxes have IoUs with ground truth boxes larger than 0.5 and the predicted label is accurate.

\subsection{Implementation Details}\label{sub:Imple}
We employ a faster R-CNN \cite{ren2015faster} with the ResNet-50-FPN \cite{lin2017feature} backbone pretrained on MS-COCO as the object detector. A pretrained feature extractor with the ResNet-50 backbone is adopted to extract image features. Both the human and object streams contains a residual convolution block and 2 FC layers. These feature extraction blocks are used to further extract instance-level features from image features according to the proposed boxes produced by the detector. The spatial stream consists of 3 convolutional layers and a max-pooling layer.  

During training, we keep the ResNet-50 backbone frozen and employ SGD optimizer. We set minibatch size as 4, an initial learning rate as 0.001 and decay by 0.6 every 10 epochs. We regard each HOI pair as positive sample if it is labeled with ground truth interaction(s), and as a negative sample otherwise. The iteration time is set to 2 since we didn't find obvious improvement when we increased it, and $\lambda$ is set to 6. We train our model for 40 epochs. Our experiments are conducted on a single Nvidia Titan X GPU.

\subsection{Ablation Studies}\label{sec:abla}
We conduct the mAP evaluation following the official setting of each dataset. For V-COCO, we evaluate the ${AP}_{role}$ for 24 HOI actions with the role setting. For HICO-DET, we conduct two modes, Default mode and Known Object mode. More details can be seen in \cite{gupta2015visual} and \cite{chao2018learning}. \whd{In this subsection, we analyze the \whe{significance of the heterogeneous nodes graph, the intra-class and inter-class message, and contextual graph attention.}} Table \ref{tab:Abla}, \ref{tab:homo} and \ref{tab:scene} show the detailed results of our ablation studies. 

\noindent\textbf{Heterogeneous Graph Structure. }
The heterogeneous graph model is the key structure for aggregating the messages which are not all equally the same in HOI and deeper inferring the relationships. To show the impact of our method, we conduct experiment on a baseline model consisting of the object detector, the feature extraction backbone and the multistream module, which are completely the same as the \whe{proposed model} has except for the absence of the most important heterogeneous graph with the learned intra-class, inter-class knowledge. 
We further separately evaluate the effectiveness of the aggregated knowledge of intra-class message and inter-class message by only computing $\mathbf{M}^{intra}$ and $\mathbf{M}^{inter}$ in the reasoning procedure, respectively. 

\setlength{\tabcolsep}{3pt}
\begin{table}[thp]
    \centering
    \caption{Ablation study results on V-COCO and HICO-DET((\%)}
    \begin{tabular}{c|ccc}
    \hline
    \hline
    \multirow{2}{*}{Method} & V-COCO & \multicolumn{2}{c}{HICO-DET} \\
        & $\mathbf{AP}_{role}$ & Default(Full) & Known Object(Full) \\
    \hline
    baseline &47.4 &13.73 &16.92 \\
    baseline + $\mathbf{M}^{intra}$ &52.0 &16.70 &20.07 \\
    baseline + $\mathbf{M}^{inter}$ &51.4 &16.13 &19.34 \\
    baseline + $\mathbf{M}^{intra}$ + $\mathbf{M}^{inter}$(proposed) &52.7 &17.57 &21.00 \\
    \hline
    w/o graph attention &49.5 &15.34 &19.86 \\
    w/o intra-class context &51.2 &16.60 &20.08 \\
    w/o interactiveness weight &51.9 &16.49 &19.95 \\
    \hline
    \end{tabular}
    \label{tab:Abla}
\end{table}

\whd{Table \ref{tab:Abla} clearly shows that both the aggregations of intra-class and inter-class messages bring \whe{considerable} improvements. \whe{With the messages, the performances increase 5.3 mAP in V-COCO and nearly 4 mAP in HICO-DET with both the two setting.} This result verifies that both the two kinds of contextual knowledge are very valuable in improving the expression of the features for HOI.} 

\noindent\textbf{Graph Attention. }
\whd{The graph attention employed in message passing layers enhances the efficiency of nodes gathering useful information from neighbors with which they have significant relations. To verify the effectiveness of the graph attention method,} we conduct experiments on the graph model that without computing the intra-class context (Eq. (\ref{eq:alpha})) and the interactiveness weight $\mathbf{w}$. \whd{In this condition, the nodes gather messages indiscriminately from the neighborhood without any guidance of the context.} 
Table \ref{tab:Abla} illustrates that compared with \whe{the proposed model, the mAP clearly decreases when we do not adopt the contextual graph attention, dropping approximately 3 mAP in V-COCO and 2 mAP in HICO-DET.} The reason behind this result is likely that without the guidance of graph contextual attention, nodes cannot selectively learn valuable information and from neighbors that resemble or have strong relevance with them and instead gather more useless messages from those unrelated nodes. 

\whe{We further evaluate the importance of intra-class context weight and interactiveness weight, respectively. Employing any of them, the performances clearly improve. By analyzing the results, we can see that it is important to realize the subordination and discriminate the messages that are not equally the same.} 

\noindent \wha{\textbf{Comparison of Heterogeneous Graph and Homogeneous Graph. }
To further verify the importance of heterogeneous structure, we design our homogeneous graph which denotes people and objects as the same nodes. In the homogeneous graph, the message passing process shares the same layer and formula. We separately set the message formulation as the same with intra-class function (Eq. (\ref{eq:homo_2})) and inter-class function (Eq. (\ref{eq:diver_2})) in heterogeneous graph. We report the results in Table \ref{tab:homo}, and show that the heterogeneous structure performs clearly better than the homogeneous. The result verifies that with heterogeneous structure, our model can learn HOIs better. }

\setlength{\tabcolsep}{4pt}
\begin{table}[htp]
    \centering
    \caption{Performance comparison of heterogeneous graph and homogeneous graph(\%) }
    \small
    \begin{tabular}{c|c|cc}
    \hline
    \hline
    \multirow{2}{*}{Method} & V-COCO & \multicolumn{2}{c}{HICO-DET} \\
        & $\mathbf{AP}_{role}$ & Default(Full) & Known Object(Full) \\
    \hline
    homogeneous(intra-class) &49.3 &13.5 &16.8 \\
    \hline
    homogeneous(inter-class) &48.9 &12.6 &16.1 \\
    \hline
    heterogeneous &52.7 &17.5 &21.0 \\
    \hline
    \end{tabular}
    \label{tab:homo}
\end{table}

\setlength{\tabcolsep}{4pt}
\begin{table}[htp]
    \centering
    \caption{The performance in different scenes on V-COCO and HICO-DET(\%). ``C" and ``S" represent complex scene and simple scene respectively }
    \begin{tabular}{c|cc|cc}
    \hline
    \hline
    Dataset & \multicolumn{2}{c}{V-COCO} & \multicolumn{2}{c}{HICO-DET} \\
    \hline
    subset & C & S & C & S \\
    \hline
    baseline &50.4 &41.2 &14.7 &12.5 \\
    \hline
    proposed(w/o $\mathbf{M}^{intra}$) &53.5 &44.1 &16.4 &15.7 \\
    \hline
    proposed(w/o $\mathbf{M}^{inter}$) &54.7 &42.4 &17.5 &14.0 \\
    \hline
    proposed &55.1 &44.1 &18.3 &15.7 \\
    \hline
    \end{tabular}
    \label{tab:scene}
\end{table}

\noindent \wha{\textbf{Analysis in Scenes with Different Complexities. }
Since our graph learns the intra-class message which is considered between instances that play the same role in HOI, the effectiveness is related to the complexity of the scenes. To analyze the affect of scenes complexity, we divide the dataset into two subsets: the complex scenes that have many entities and multiple interactions, and the simple scenes that only have one person and one object. According to the labels, We count about 4k complex scenes and about 600 simple scenes in V-COCO test set, and about 7k complex scenes and 2.5k simple scenes in HICO-DET test set. We evaluate the heterogeneous graph on the two subsets and make comparisons with the baseline model to reveal the effectiveness. Furthermore, we also evaluate the intra-class message and the inter-class message. Table \ref{tab:scene} illustrates the comparison results, which show that our model makes great progress in both complex and simple scenes. \whb{The intra-class messages explicitly improve the result in complex scenes, which reveals that by reasoning among different nodes with intra-class context, our model can tell which homogeneous entities are relevant and learn more from relevant entities, while eliminate irrelevant entities. Therefore, the model can selectively propagate useful messages between homogeneous nodes and improve the recognition.} The mAP increases with the inter-class messages, which shows that the inter-class messages can lead to better understanding between subject and object. }

\subsection{Performance and Comparison}
This section we compare the performances of the heterogeneous graph model with previous deep methods. For HICO-DET we further evaluate the results with the Full, Rare and Non-Rare settings, which consider 600, 138, and 462 HOI actions, respectively. \whe{Table \ref{tab:VCOCO} and Table \ref{tab:HICO-DET} illustrate the performances on V-COCO and HICO-DET. } 

\setlength{\tabcolsep}{4pt}
\begin{table}[htbp]
    \centering
    \caption{Performance comparison on the V-COCO}
    \begin{tabular}{c|c}
        \hline
        \hline
        Method & ${AP}_{role}$(\%) \\
        \hline
        Gupta.S \cite{gupta2015visual} &31.8 \\
        InteractNet \cite{gkioxari2018detecting} &40.0 \\
        GPNN \cite{qi2018learning} &44.0 \\
        iCAN \cite{gao2018ican} &45.3 \\
        Feng \emph{et al.} \cite{feng2019turbo} &42.0 \\
        XU \emph{et al.} \cite{Xu_2019_CVPR} &45.9 \\
        Wang \emph{et al.} \cite{wang2019deep} &47.3 \\
        Prior \cite{li2018transferable} &47.8 \\
        RPNN \cite{zhou2019relation} &47.5 \\
        PMFNet \cite{wan2019pose} &52.0 \\
        proposed &\textbf{52.7} \\
        \hline
    \end{tabular}
    \label{tab:VCOCO}
\end{table}

\setlength{\tabcolsep}{4pt}
\begin{table}[htbp]
    \centering
    \caption{Performance comparison on the HICO-DET}
    \begin{tabular}{c|cccccc}
    \hline
    \hline
    \multirow{2}{*}{Method} & \multicolumn{3}{c}{Default}(\%) & \multicolumn{3}{c}{Known Object}(\%) \\
        & Full & Rare & Non-Rare & Full & Rare & Non-Rare \\
    \hline
    HO-RCNN \cite{chao2018learning} &7.81 &5.37 &8.54 &10.41 &8.94 &10.85 \\
    InteractNet \cite{gkioxari2018detecting} &9.94 &7.16 &10.77 &- &- &- \\
    GPNN \cite{qi2018learning} &13.11 &9.34 &14.23 &- &- &- \\
    iCAN \cite{gao2018ican} &14.84 &10.45 &16.15 &16.26 &11.33 &17.73 \\
    Feng \emph{et al.} \cite{feng2019turbo} &11.4 &7.3 &12.6 &- &- &- \\
    XU \emph{et al.} \cite{Xu_2019_CVPR} &14.70 &13.26 &15.13 &- &- &- \\
    Wang \emph{et al.} \cite{wang2019deep} &16.24 &11.16 &17.75 &17.73 &12.78 &19.21 \\
    Prior \cite{li2018transferable} &17.03 &13.42 &\textbf{18.11} &19.17 &15.51 &\textbf{20.26} \\
    No-frills \cite{gupta2019no} &17.18 &12.17 &18.68 &- &- &- \\
    RPNN \cite{zhou2019relation} &17.35 &12.78 &18.71 &- &- &- \\
    PMFNet \cite{wan2019pose} &17.46 &15.65 &18.00 &20.34 &17.47 &21.20 \\
    proposed &\textbf{17.57} &\textbf{16.85} &17.78 &\textbf{21.00} &\textbf{20.74} &\textbf{21.08} \\
    \hline
    \end{tabular}
    \label{tab:HICO-DET}
\end{table}

For \textbf{V-COCO}, the proposed model outperforms state-of-the-art methods and improves the mAP by 0.7. \whe{We outperform previous graph network GPNN by 8.7 mAP, demonstrating the effectiveness of the heterogeneous graph and the importance to discriminate intra-class message and inter-class message.} 

For \textbf{HICO-DET}, the results also demonstrate the significance of our method. We achieve state-of-the-art performance in both Default mode and Known Object mode. Compared with GPNN, our method boosts the performance 34\% in Default mode.

\begin{figure}[ht]
    \centering
    \includegraphics[width=\linewidth]{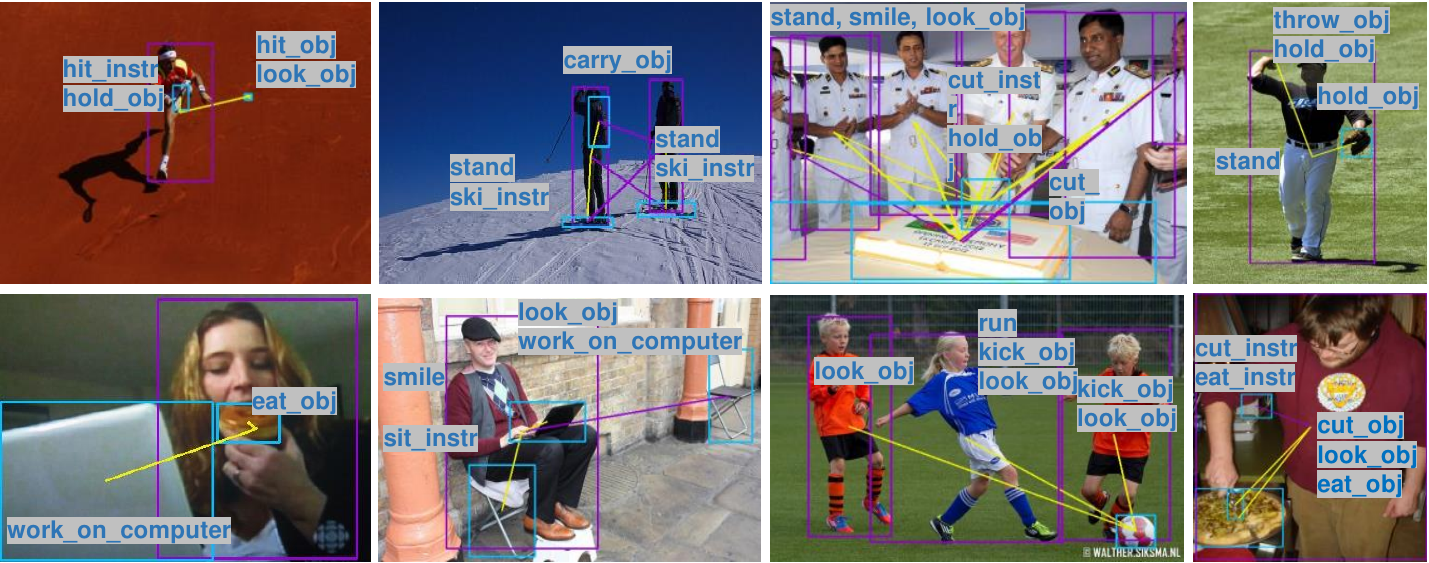} 
    \caption{Visual samples of HOI detection in VCOCO. Human and object instances are denoted by purple and blue bounding boxes. The yellow lines represent that the linked instances have interactions with each other. The figures show some more complex cases that with multiple different interactions in the images. }
    \label{visualize2}
\end{figure}

\begin{figure}[t]
    \centering
    \includegraphics[width=\linewidth]{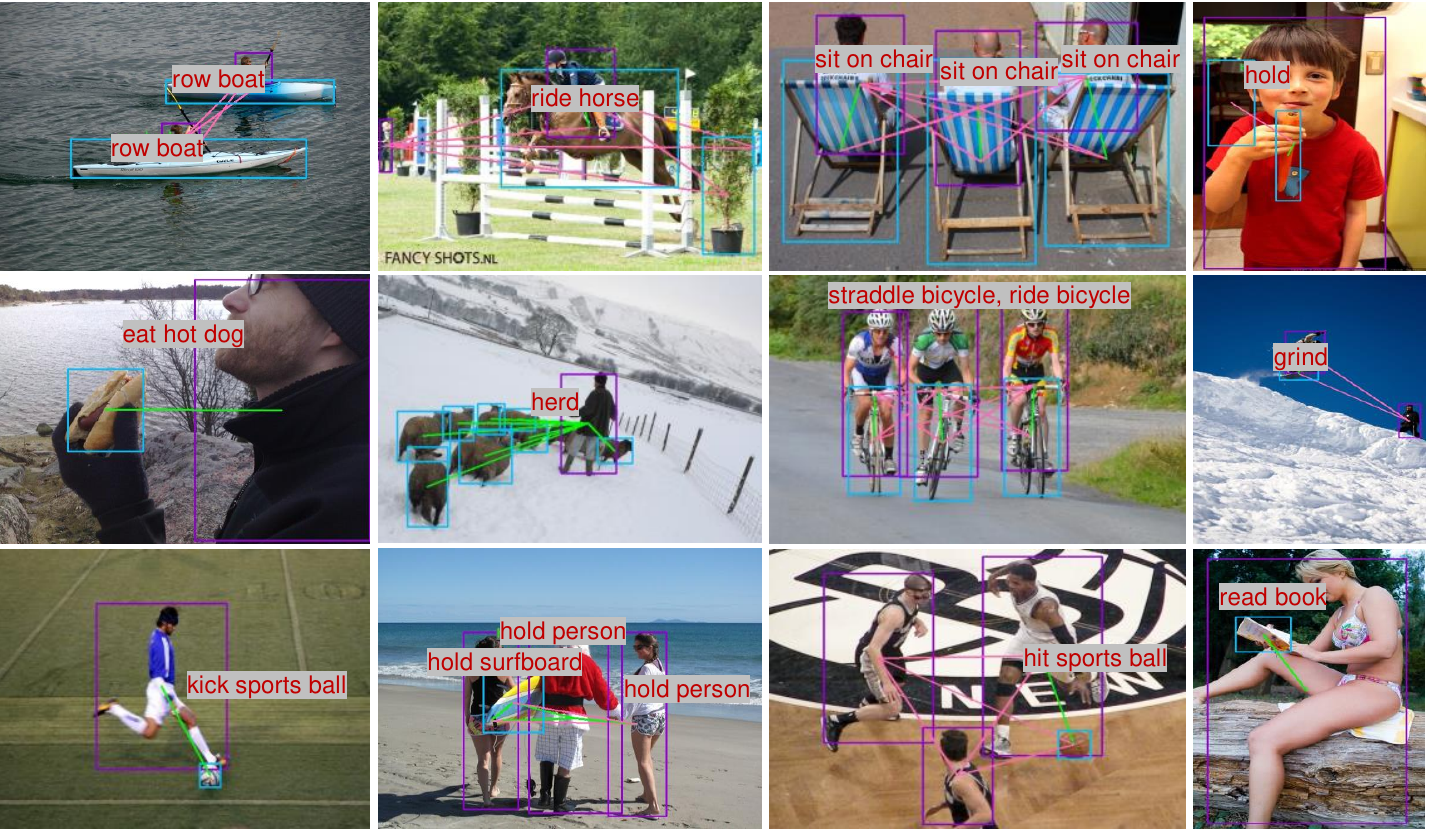} 
    \caption{Visual samples of HOI detection in HICO-DET. Human and object instances are denoted by purple and blue bounding boxes. The green lines represent that the linked instances have interactions with each other, while the pink lines mean no interaction within the instances.}
    \label{visualize1}
\end{figure}

Fig. \ref{visualize2} and Fig. \ref{visualize1} illustrate some visualized HOI detection samples tackled by our model in V-COCO and HICO-DET separately. These samples show the effectiveness of our proposed method in detecting HOIs in activity scenes with different complexities varying from the simple kind with only one human-object pair to the complicated kind with multiple entities and interactions in the image. The samples also shows that the model works well at recognizing whether the subjects and objects are interactive in different scenes, which reveals the effect of the inter-class context. 

\section{Conclusions}
Since the messages between humans and objects playing the same roles (subjects or objects) and that playing different roles (subject-object) are not equally the same in human-object interactions, in this paper We explore the unequal relations and messages by proposing a heterogeneous graph model. The graph discriminates the messages as intra-class message and inter-class message, and adopt specialized methods and graph attention to aggregate them. We find that the heterogeneous graph further exploring the messages that are not equally the same is effective in reasoning HOI. In V-COCO and HICO-DET datasets, the proposed model gains promised results and outperforms the state-of-the-art method.

\textbf{Acknowledgements.}
This work was supported partially by the National Key Research and Development Program of China (2018YFB1004903), NSFC (U1911401,U1811461), Guangdong Province Science and Technology Innovation Leading Talents (2016TX03X157), Guangdong NSF Project (No. 2018B030312002), Guangzhou Research Project (201902010037), and Research Projects of Zhejiang Lab (No. 2019KD0AB03).

%
%
\bibliographystyle{splncs04}
\bibliography{egbib}
\end{document}